\documentclass[twoside,twocolumn]{article}
\usepackage{natbib,pifont,multirow,longtable}
\usepackage{supertabular}
\usepackage{graphicx,color,amssymb,amsmath}
\usepackage{bbm,bm}
\usepackage{authblk}

\usepackage{subfig}

\usepackage{pgfplots}

\usepackage[draft]{hyperref}

\usepackage{todonotes}

\newcommand{\argmax}[1]{ \underset{#1}{\operatorname{arg\,max}} }

\newcommand{\vbar}{\,|\, }

\newif\ifexclude

\title{Local and Global Trend Bayesian Exponential Smoothing Models}

\author[1]{Slawek Smyl}
\author[2,3,*]{Christoph Bergmeir}
\author[4]{Alexander Dokumentov}
\author[3]{Xueying Long}
\author[3]{Erwin Wibowo}
\author[3]{Daniel Schmidt}

\affil[1]{Walmart Labs}
\affil[2]{Department of Computer Science and Artificial Intelligence\\
	University of Granada, Spain.}
\affil[3]{Department of Data Science and Artificial Intelligence\\
    Monash University, Australia.}
\affil[4]{Let's Forecast\\
\url{https://letsforecast.com/}	}
\affil[*]{Corresponding Author: bergmeir@ugr.es}
	
\date{}

\begin{document}
\maketitle

\begin{abstract}
This paper describes a family of seasonal and non-seasonal time series models that can be viewed as generalisations of additive and multiplicative exponential smoothing models, to model series that grow faster than linear but slower than exponential. Their development is motivated by fast-growing, volatile time series. 
In particular, our models have a global trend that can smoothly change from additive to multiplicative, and is combined with a linear local trend. Seasonality when used is multiplicative in our models, and the error is always additive but is heteroscedastic and can grow through a parameter sigma.
We leverage state-of-the-art Bayesian fitting techniques to accurately fit these models that are more complex and flexible than standard exponential smoothing models. 
When applied to the M3 competition data set, our models outperform the best algorithms in the competition as well as other benchmarks, thus achieving to the best of our knowledge the best results of per-series univariate methods on this dataset in the literature. An open-source software package of our method is available.

\end{abstract}

\section{Introduction}\label{sec:intro}

Despite being introduced over 50 years ago, exponential smoothing methods remain among the most widely used methods for forecasting~\citep{Goodwin2010Holt}. Their enduring popularity stems primarily from their relative simplicity, robustness, flexibility, and good forecasting performance~\citep{GardnerJr.2006Exponential}. For example, exponential smoothing models were able to outperform several more sophisticated and complex algorithms in the influential M3 forecasting competition \citep{Makridakis2000M3}, and exponential smoothing models were successfully used as building blocks in many of the methods used in the recent M4 competition~\citep{Makridakis2018M4}.

Exponential smoothing was originally created as a simple heuristic, in which the next point forecast is given by the previous point forecast plus a correction term that is proportional to the error made by the previous forecast. Over the years, the basic exponential smoothing technique has been extended to include additional aspects such as seasonality (additive and multiplicative), local trends (additive, multiplicative, and damped versions of both), and more complex stochastic error term specification (additive and multiplicative). Exponential smoothing techniques are frequently referred to in the literature by the initialism ``ETS'', which may stand for either ExponenTial Smoothing, or Error, Trend, and Seasonality. \cite{Hyndman2002state} and \cite{Hyndman2008Forecasting} systematically consolidated the various exponential smoothing models, and gave them a solid theoretical base, by reformulating them within a state-space framework with a single source of error. Subsequently, this theoretical contribution was followed by the creation of the \emph{forecast} package \citep{Hyndman2008Automatic} in the R programming language~\citep{RCT2018R}, which implements ETS models in an accessible and user-friendly way within an open-source framework. The package has since become the {\em de facto} standard tool for the forecasting of time series in many areas. A more recent continuation of this development is via the \emph{fable} package \citep{ohara2021fable}, which reimplements and extends the functionality of the \emph{forecast} package.
Apart from these main lines of research, other authors have developed various robust versions of exponential smoothing, e.g., \citep{Cipra1992robust, Gelper2010Robust, crevits2016forecasting, crevits2018robets}. \cite{Svetunkov2022Complex} developed the complex exponential smoothing method, and \cite{sbrana2020forecasting} present a damped trend model that uses a structural approach, i.e., a multiple sources of error formulation.

The ETS models that are currently in use (as described in \citet{Hyndman2008Forecasting}, and implemented in the \emph{forecast} package), are either linear or exponential in their trend, with the only option currently available to potentially bridge this gap being a damped trend. So, the choice is limited especially if the underlying growth rate cannot be suitably modeled by these available options. 
One way to currently address this shortcoming is by using data transformations as a preprocessing step, such as a logarithm or Box-Cox transformation \citep{Box1964analysis}. However, the logarithm is a very strong transformation that can lead to explosive behaviour. The strength of the Box-Cox transformation can be adjusted using its parameter $\lambda$, but choosing this parameter is not trivial and we observe in the experiments of our ablation study that standard methods for this task also often lead to explosive forecasts.
Attempts have been made in adjacent fields to develop more flexible trend formulations, for example, \cite{parzen1982ararma} presents a model where the differencing of ARIMA is replaced by other, slightly more complicated non-stationary AR models, thus allowing for more flexible trend choices. 
Apart from the relatively unflexible trend, the ETS models assume normality of the errors, homoscedasticity of the errors for additive models, and have some further shortcomings described in details in Section \ref{sec:short}.

\subsection{Bayesian ETS}

The majority of development in the ETS space has been through the lense of frequentist statistical approaches. In contrast, the application of Bayesian methods to exponential smoothing has been comparatively less studied. Important examples of Bayesian ETS approaches being proposed in the works of \cite{Andrawis2009new} and \cite{Bermudez2010Bayesian}. \cite{Corberan-Vallet2013Bayesian} subsequently proposed a related method for time series containing zeros,
\cite{Bermudez2009Multivariate} presented a model suitable for multivariate forecasting, and \cite{corberan2011forecasting} extended the approach to forecasting of time series with correlated random disturbances.
The work of \cite{Bermudez2010Bayesian} examined the topic of Bayesian estimation of basic ETS models in a holistic way. The performance of resulting model was comparable to the best forecasting models in the M3-competition, as well as forecasts produced from popular automatic packages \citep{Bermudez2010Bayesian}. An interesting result of that study was the demonstration that the use of Bayesian estimation meant the resulting model was able to produce accurate prediction intervals, which were found to be close to the empirical intervals. This is an important feature that the classical exponential model from the M3 competition is lacking. More recently, \cite{tsionas2021bayesian} extended the work of \cite{sbrana2020forecasting} to a Bayesian model with a multiple source of error formulation.
Interestingly, despite the promising results emerging from the works in this area, the Bayesian approach has not found widespread adoption in the application of ETS models. This could be due to it being relatively slow or its relatively complicated implementation and a lack of readily available software in this space.

\subsection{A motivating example}

\begin{figure*} 
  \centering
  \subfloat[Original series]{\includegraphics[width=0.5\textwidth]{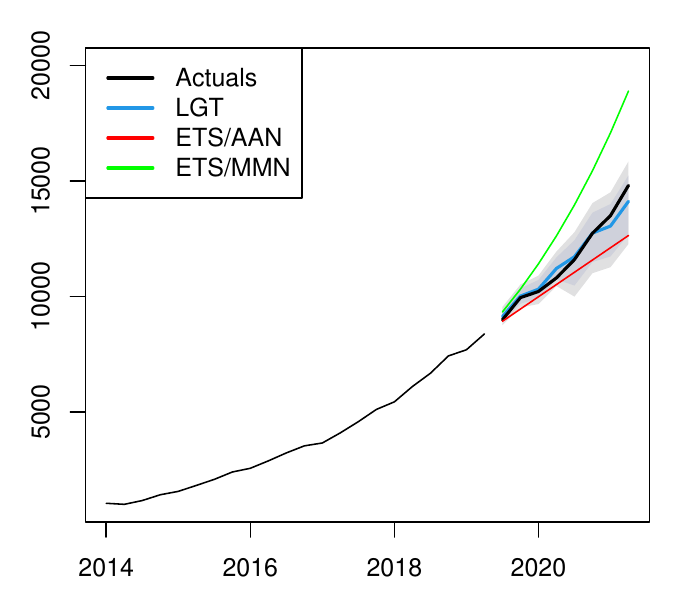}}
  \subfloat[Log of series]{\includegraphics[width=0.5\textwidth]{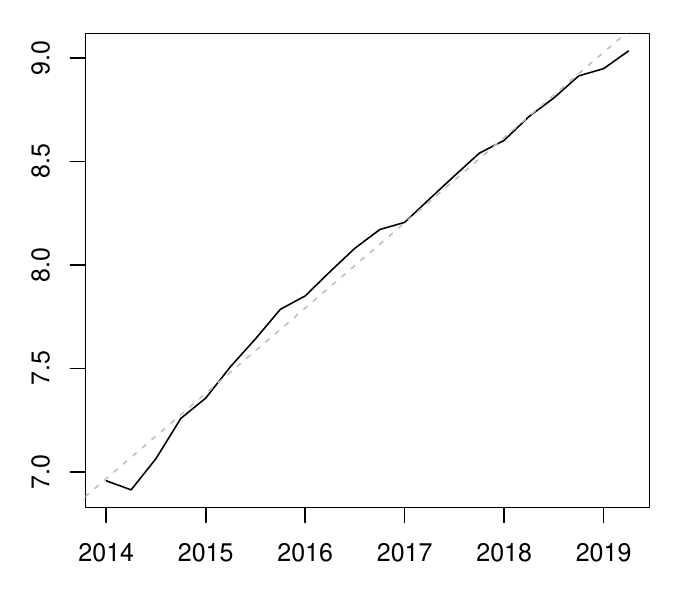}}
  \hfill
  \subfloat[Absolute differences of the series]{\includegraphics[width=0.5\textwidth]{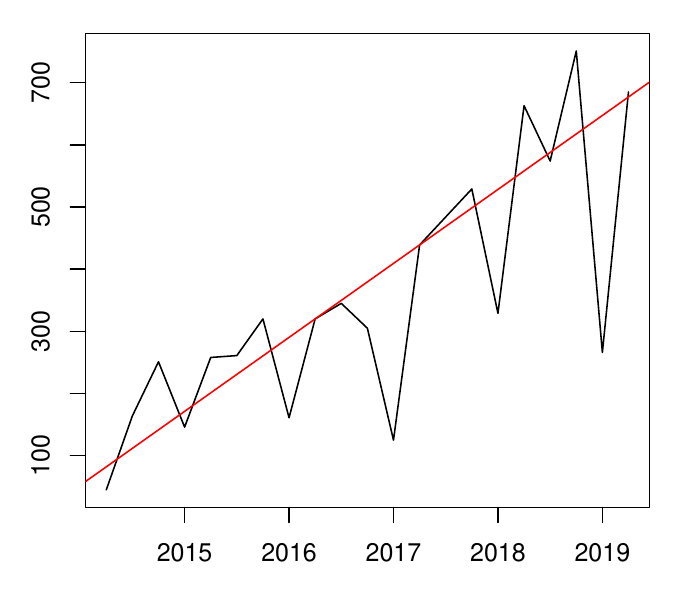}}
  \subfloat[Absolute percentage differences of the series]{\includegraphics[width=0.5\textwidth]{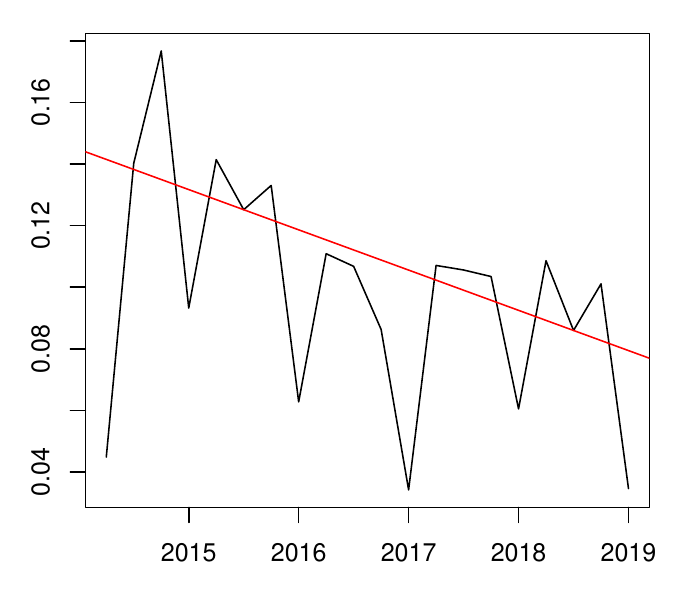}}

  \caption{a) Quarterly revenue of Amazon Web Services (in million USD, source: statista.com). Forecasts from ETS with linear and exponential trends show that the trend is either under- or over-estimated. LGT is able to capture the trend adequately. b) The logarithm of the series confirms that the series grows slower than exponential (as the log of the series doesn't grow linearly). c) Absolute differences of original data. Assuming errors are bigger than changes in trend, these values estimate absolute errors. Under this assumption, the graph shows that errors grow over time. d) Absolute differences divided by first value, i.e., $\frac{|y_{t+1} - y_t|}{y_t}$. We see that errors are not proportional to the trend.}
  \label{fig:aws}
\end{figure*}

There are many practical applications, e.g., in cloud infrastructure, rideshare, and other tech businesses, where characteristics of the ETS framework mentioned above do not adequately capture the time series in consideration. 
In many applications, time series grow faster than linear but slower than exponential, and in situations with volatile time series, the Gaussian error assumption is violated. 
Strong trends will in the real world eventually slow, e.g., in product lifecycles. This has led to the development of damped trend models in the literature. 
However, we argue that there are many business use cases where this de-acceleration of the trend may still be far away in the future, and unclear when it will happen. Large technological trends, growth in technology governed by Moore's Law, or growth of a highly successful startup company may lead to strongly trending series over decades, and even when they slow down they may still show a growth that is faster than linear. The motivation for our method is to produce short- and mid-term forecasts for such series.

As an illustrative example, we use in the following the quarterly revenue of Amazon Web Services, a cloud infrastructure provider. Figure~\ref{fig:aws} shows the series, and indications that it grows faster than linear but slower than exponential. We also show forecasts of ETS with a linear and exponential trend, and our proposed method LGT which is able to forecast the series better. We furthermore can see that the error in the series grows, but at a slower pace than proportional to the level. Thus, expressing scale of the error as a function of level is important in time series with error levels that are one or two orders of magnitude smaller at the beginning than at the end of the series, so that homoscedastic models are unsuitable in this case. At the same time, while volatility grows with time series levels, it is seldom proportional to the level.

\subsection{Our Contributions}

In this paper we propose a number of practically motivated extensions to the ETS family of models with additive error and additive trend, to convert them into flexible nonlinear models. As the models are likely to be analytically unsolvable, and our models use non-Gaussian error distributions, we fit them using a probabilistic programming tool~\citep{StanDevTeam2015Stan} that employs a fast Markov Chain Monte Carlo engine~\citep{Hoffman2014no}. 
In sum, the main contributions of our paper are as follows:

\begin{enumerate}
\item Motivated directly by phenomena commonly observed in real-world time series, we implement a number of extensions of the ETS
framework that are designed to increase the flexibility of the base models. In particular, we introduce a novel global trend modelling that can be flexibly adjusted to growth rates that are between linear and exponential. 
\item By using Bayesian model fitting procedures, our more flexible models can still be fitted accurately. We can relax the usual Gaussian assumptions and still obtain accurate prediction intervals, even for small sample sizes. We implement both a standard version of the sampler, rendering the implementation extensible and easy to understand, and a bespoke specialised Gibbs sampler that is roughly an order of magnitude faster than the standard implementation.
\item This gives us in sum an acceptably fast model that is able to outperform even the most competitive benchmarks in this research area in terms of accuracy. We provide documented and tested open-source software implementations of our proposed methods.
\end{enumerate}

The methods presented in this paper are available in the \emph{Rlgt} package on CRAN.\footnote{\url{https://cran.r-project.org/package=Rlgt}} The \emph{Rlgt} package is a comprehensive implemention of models discussed in this paper, together with further documentation and illustrations regarding their use.

The remainder of this paper is structured as follows: Section~\ref{sec:ets} discusses ETS models in detail, Section~\ref{sec:short} describes the shortcomings of classical ETS models, and Section~\ref{sec:bayes} discusses Bayesian model fitting and MCMC sampling. Sections~\ref{sec:non-seas} and \ref{sec:seas} present the non-seasonal and seasonal versions of our new model  respectively, Section~\ref{sec:performance} describes the experiments and results, 
and Section~\ref{sec:conclusions} concludes the paper.

\section{Exponential Smoothing}\label{sec:ets}

The exponential smoothing algorithm is fundamentally based on the concept of time series
decomposition. As discussed in Section~\ref{sec:intro}, ETS performs a decomposition of the time series data into the three main components: (i) stochastic error; (ii) a trend; and (iii) a seasonal component.
The trend represents the long-term increase and decrease in the time series, and is not necessarily linear over time. The seasonal component represents the perturbation in the time series due to seasonal factors, e.g., quarter of the year or day of the week. Seasonality patterns last for a fixed period of time. Finally, the error component models the random components which cannot be explained by the model. 

These components can interact with each other in either additive or multiplicative ways. Other interactions are also possible in the ETS model, such as damped trends, in which the strength of the trend reduces over time \citep{Gardner1985Exponential}.
Exponential smoothing methods were proposed as a simple way of generating fast and
reliable forecasts for a wide range of time series data in various application areas that do not make too many strong, parametric assumptions.

\subsection{ETS model family nomenclature}
A wide range of different ETS-based models have been introduced over the years. Following the nomenclature in ~\cite{Hyndman2018Forecasting}, ETS variants commonly model the trend as either additive (A), additive damped (Ad), multiplicative (M), or as having no trend (N). Seasonality is usually modelled as being either additive (A), multiplicative (M), or not present (N), and the error term is usually either additive (A) or multiplicative (M) in nature.
The ETS model family constitutes the set of models that can be built from all possible combinations of these choices for error term, trend, and seasonality. For example, an ETS model with additive error, additive damped trend and multiplicative seasonality, which is commonly known as the Holt-Winters' damped method, would be denote as an ``AAdM model'' under the common nomenclature. Furthermore, if a component is not further specified, a ``Z'' is used as a wildcard placeholder for the missing term; for example, an AAZ model is a model with additive error, additive trend, and any form of seasonality.

\subsection{Classical ETS model estimation}

The classical ETS models are usually fitted via the method of maximum likelihood. This technique is the most popular technique for estimating model parameters in classical statistics. This popularity is primarily due to the strong asymptotic properties the maximum likelihood estimates are known to possess (under suitable regularity conditions), such as consistency, asymptotic efficiency, and asymptotic normality \citep{Fahrmeir1985Consistency}.
The method of maximum likelihood is based on the notion of maximising the joint likelihood of the data under the model
\[
    p(y_1,\ldots,y_n \vbar \bm{\theta}) \equiv p({\bf y} \vbar \bm{\theta}).
\]
Here, $\bm{\theta}$ is a vector of model parameters and ${\bf y}$ is a vector of $n$ observed values, i.e., the likelihood function $p({\bf y} \vbar \bm{\theta})$ is defined as the probability of generating a certain set of data ${\bf y}$ given the parameter values $\bm{\theta}$. The estimates of the parameters are defined as the values of the parameters which maximise the likelihood function, i.e.,
\begin{equation}
\label{equ:ll1}
\hat{\bm{\theta}} = \argmax{\bm{\theta}\in\Theta} \left\{ p \left( \bm{\theta} \vbar {\bf y} \right) \right\}
\end{equation}
Here, $\hat{\bm{\theta}}$ denotes the maximum likelihood estimate (MLE) and $\Theta$ denotes the valid parameter space. For the classical ETS models, the parameter vector $\bm{\theta}$ usually consists of (a mix of) the following parameters: a damping coefficient, an error variance, initial trend terms, initial seasonality
factors, and smoothing factors for the level, trend, and seasonality. For the purpose of forecasting, it is usual that the estimated parameters are plugged into the relevant equations to produce future predictions~\citep{Hyndman2018Forecasting}.

\subsection{Non-seasonal (damped) trend model}
\label{sec:AAN}

One particular model from the ETS model family, the non-seasonal (damped) linear trend model (AAdN), is given by
\begin{align}
\label{equ:aan1} y_{t+1} &\sim N \left(\hat{y}_{t+1}, \sigma^2 \right),\\
\label{equ:aan2} \hat{y}_{t+1} &= l_{t} + \phi b_{t},\\
\label{equ:aan3} l_{t+1} &= \alpha y_{t+1}+ \left( 1- \alpha  \right) \left(l_{t} + \phi b_t\right),\\
\label{equ:aan4} b_{t+1} &= \beta  \left( l_{t+1}-l_{t} \right) + \left( 1- \beta  \right) \phi b_{t},
\end{align}
where $y_{t}$ denotes the value of the dependent variable of interest at time $t$, $\hat{y}_{t+1}$ is the conditional expectation of $y_{t+1}$ given the information up to time $t$, $\sigma$ denotes the standard deviation of the error distribution, $l_{t}$ denotes the level at time $t$, and $b_{t}$ the local trend at time $t$. The parameters to be fitted are $\alpha \in (0,1)$, a smoothing parameter for the level term; $\beta \in (0,1)$, a smoothing parameter for the local trend term; $\phi \in (0,1)$, a damping factor; and initial values for level, $l_{1}$, and slope, $b_{1}$.

Equation~(\ref{equ:aan1}) models the distribution of $y_{t+1}$; specifically, it assumes that the value of $y_{t+1}$ is distributed as per a normal distribution with mean $\hat{y}_{t+1}$ and standard deviation $\sigma$. Equation~(\ref{equ:aan2}) states that the one-step-ahead prediction value $\hat{y}_{t+1}$ is formed as the sum of the previous level and trend values. Equation~(\ref{equ:aan3}) describes the evolution of the level over time; the estimate of the level at time $t+1$ is given by the weighted average of the observed value $y_{t+1}$ and the previous estimate of the level $l_t$. The initial level value $l_1$ is treated as a parameter to be estimated from the data.

Equation~(\ref{equ:aan4}) describes the evolution of the trend across the time horizon. Similar to Equation~(\ref{equ:aan3}), the updating process of the trend is based on the weighted average of the current trend value $\left(l_{t+1}-l_{t}\right)$  and the previous estimate of the trend $b_t$.
As with the level, the initial value $b_{1}$ is treated as a parameter and estimated. The parameter $\phi$ is a damping factor that reduces the influence of the trend; if $\phi=1$ the model reduces to the usual AAN (i.e., undamped) model.

We note that when using a linear trend, we are making the assumption that we expect that in addition to the level changing over time, we also expect that the trend will change over time. The weighted average estimate for the trend value usually works best if the trend changes gradually over time in the corresponding time series data. More comprehensive explanations of this model can be found in~\cite{Hyndman2018Forecasting}.

\subsection{Seasonal (damped) trend model}
\label{sec:AAM}

The seasonal counterpart of the non-seasonal (damped) linear trend model introduces a multiplicative seasonal pattern to account for potential seasonality. This model is also known as the Holt-Winters’ multiplicative method \citep{Hyndman2018Forecasting}. 
\begin{align}
	\label{equ:aas1} y_{t+1} &\sim N \left(\hat{y}_{t+1}, \sigma^2\right)\\
	\label{equ:aas2} \hat{y}_{t+1} &= (l_{t} + \phi b_{t}) s_{t+1-m},\\
	\label{equ:aas3} l_{t+1} &= \alpha \left( \frac{y_{t+1}}{s_{t+1-m}} \right) + \left( 1- \alpha  \right) \left(l_{t} + \phi b_t\right),\\
	\label{equ:aas4} b_{t+1} &= \beta  \left( l_{t+1}-l_{t} \right) + \left( 1- \beta  \right) \phi b_{t},\\
	\label{equ:aas5} s_{t} &= \zeta \left( \frac{y_{t}}{l_{t}} \right) + (1 - \zeta) s_{t-m}.
\end{align}
This extension introduces three new quantities: the $s_t$ represents the seasonal factor prevailing at time $t$, $\zeta \in (0,1)$ represents a smoothing parameter for the seasonal factors, and $m>1$ represents the number of seasons comprising a complete period. The main differences between the AAdN model described in Section \ref{sec:AAN} and the seasonal extension are:

\begin{itemize}
\item In comparison to Equation (\ref{equ:aan2}), Equation (\ref{equ:aas2}) introduces a multiplicative factor of $s_{t+1-m}$ which estimates the prevailing seasonal factor at time $t+1$. This factor is estimated $m$ periods before (at this point the seasonal factor is not yet known, so the best estimate that we have is the seasonal factor obtained at time $t+1-m$).
\item The updating process of the level in Equation (\ref{equ:aas3}) is based on the weighted average of the current level value (note that the data value is modeled as a product of the level and seasonality factor, so the level value is given by $y_{t+1}/s_{t+1-m}$) and the one-step prediction of the current level value.
\item Equation (\ref{equ:aas5}) describes the evolution of the seasonality factor. Similar to the trend and level adjustment process, the seasonality factor is estimated as a weighted average of the current seasonality value $y_t/l_t$ and the most recent estimate of the corresponding seasonality value $s_{t-m}$.
\end{itemize}

\section{Weaknesses of classical ETS models}
\label{sec:short}

As briefly discussed in Section~\ref{sec:intro}, classical ETS models have several shortcomings that can adversely affect their performance in practice. We now describe these weaknesses in more detail.

\subsection{Limited functional form of trend}
\label{sec:limited:form}

Despite a variety of different forms of trends being considered in classical ETS models, the choice may frequently still be too limited; this is illustrated by the motivating example of quarterly AWS revenue, shown in Figure~\ref{fig:aws}. Clearly, the trend appears to be growing faster than an additive form but slower than a multiplicative form. As mentioned earlier, one way currently available to address this issue is by using a transformation as a pre-processing step, such as a logarithm or a Box-Cox transformation. See, for example, \cite{beaumont2014data}. The logarithm is a very strong transform and often leads to explosive behaviour of the forecasts. The Box-Cox transformation has a parameter $\lambda$ that controls its strength, but it is not trivial to choose. Including its estimation in the MLE has the problem that different values of $\lambda$ lead to different autocorrelation structures, making it necessary to, e.g., change the ARIMA order during the search \citep{Guerrero1993Time}. The method of \cite{Guerrero1993Time} chooses $\lambda$ independently of the modelling, in a way that the seasonal variation is stabilised across the series. While in our motivating example of AWS revenue the method is able to achieve a good result similar to our proposed method, in broader experiments, e.g., on the full M3 dataset, the method often still leads to explosive behaviour of the forecasts. See the ablation study in the experimental section of our paper for details.

\subsection{Non-normal error distribution}
Statistical models based on ETS models, both additive or multiplicative, usually assume that the error terms are identically and independently distributed normal random variables. This assumption is fairly restrictive for most applications, and is not necessarily appropriate for modelling volatile time series data, such as the example from Figure~\ref{fig:aws}. The assumption of normality  implicitly assumes that the probability of obtaining a value that highly deviates from the mean (i.e., an outlier) is extremely low. In some real data, such as the movements of stock prices, there is a general consensus that the normality assumption tends to be violated due to high variance, especially in the time of crises. Despite these empirical observations, there are few works in the literature to mitigate this issue. \cite{beaumont2014data} addresses the problem with the use of a Student $t$-distribution instead of a normal distribution, and a Johnson error trend seasonal transform. However, such changes are oftentimes not straightforward in the MLE framework and therefore the normality assumption tends to be retained in classical ETS models as even small deviations from the normality assumption may greatly complicate the likelihood function; this can cause considerable issues as maximisation of the likelihood function is the key step in standard MLE-based procedures. Consequently, the assumption of normality is usually required in classical ETS models, as otherwise the MLE may be difficult to find, and other inferential quantities derived from the likelihood, such as confidence intervals may also be very difficult to obtain.

\subsection{Heteroscedastic error distribution}

In addition to being non-normally distributed, the scale of the errors in time series data is often not homoscedastic, i.e., it may vary over time. This issue has been recognised even in classical ETS models, and  various approaches based on innovation state space representations have been introduced to deal with it, with varying levels of success, see, e.g., \cite[Ch. 4, 19]{Hyndman2008Forecasting}, and \cite{beaumont2014data}. For example, the multiplicative error form in the state space models may allow the error to depend linearly on the current value of the data. Similar to the discussion of the form of the trends in Section~\ref{sec:limited:form}, the way in which the error scale is linked to the trend is usually quite limited, and may not be sufficient to accurately model more general time series data. For example, the state-space model might not be able to handle an error function which grows with the data over time (i.e., faster than an additive error form) but at a sub-linear rate (i.e., slower than the multiplicative error form).

\section{Bayesian estimation}\label{sec:bayes}

The overwhelming majority of previous work on exponential smoothing utilises frequentist inferential tools, such as maximum likelihood and confidence intervals, to estimate model parameters and provide measures of uncertainty. Maximum likelihood estimation tends to suffer from high parameter variance, particularly for small sample sizes and probability models with heavy tails, such as the $t$-distribution; similarly, due to the great difficulty in deriving exact confidence intervals for all but the simplest of models, most uncertainty quantification relies on asymptotic arguments, and may provide poor estimates of variability for small sample sizes.

In contrast to this, we use the Bayesian paradigm of statistical inference to fit our proposed ETS models. In the Bayesian approach one treats the unknown model parameters, say $\theta$, as realisations of random variables coming from some prior distribution, $\pi(\theta)$, and argues conditionally on the observed data sample, $y$, to form the posterior distribution $p\left( \theta \vbar y \right)$:
\begin{equation}
\label{eq:post}
p\left( \theta \vbar y \right) = \frac{p( y \vbar \theta ) \pi(\theta)}{p(y)},
\end{equation}
where $p(y \vbar \theta)$ denotes the likelihood function, and $p(y)$ is the marginal probability of the data
under any possible model as a normalising factor. The posterior distribution describes how likely specific values of $\theta$ are to be the unknown population value of $\theta$, in light of the data and prior beliefs. The Bayesian approach naturally handles complete parameter uncertainty through the mechanism of the prior distribution. This same mechanism also allows us to specify any prior information about the unknown model parameters in a very transparent manner, and as such tends to offer a form of regularisation that can stabilise parameter estimates. In the context of ETS, regularisation has recently been shown to be beneficial, in a frequentist approach by \cite{pritularga2023shrinkage}. An additional strength of the Bayesian approach is that given the posterior distribution over the unknown model parameters, it is, at least in principle, straightforward to generate the posterior distribution over any function of the model parameters via standard transformation of random variables techniques. This can be used, for example, to generate posterior distributions over point or distributional forecasts, which naturally incorporate the uncertainty in parameter estimates.

\subsection{Monte Carlo Markov Chain (MCMC) Methods}

Despite the attractive theoretical advantages of the Bayesian approach, the primary weakness is in the potential difficulties in implementation. Computation of the normalising factor $p(y)$ in (\ref{eq:post}) involves high-dimensional integration, and is in general intractable. Even if one could compute it, effectively exploring an arbitrary multidimensional probability distribution is problematic. Instead, one generally resorts to some form of approximation; one of the most popular approximation techniques is via simulation of values of $\theta$ from the posterior distribution. This is usually done via Monte Carlo Markov Chain (MCMC) techniques, that do not require knowledge of the normalising term, and are thus very convenient. Given a random sample of $m$ values drawn from (\ref{eq:post}), say $\theta^{(1)},\ldots,\theta^{(m)}$, it is straightforward to compute quantities such as the posterior mean, standard deviation and intervals using the corresponding empirical (sample) quantities. Furthermore, if we have some function of the model parameters,  $f(\theta)$, we can approximate the posterior distribution of this function by using
\[
f\left(\theta^{(1)}\right),\ldots,f\left(\theta^{(m)}\right),
\]
i.e., we simply evaluate $f(\cdot)$ for each sample $\theta^{(i)}$ we have drawn from the posterior distribution. In the context of this paper, the typical functions we may evaluate would be quantities such as forecasts over future data.

\subsection{Probabilistic Programming}

The application of Bayesian statistical techniques to practical data science problems has increased dramatically in the last two decades. This has primarily been driven by increases in computational power, coupled with the introduction of easy-to-use Bayesian probabilistic programming tools 
such as Stan \citep{StanDevTeam2015Stan} and JAGS \citep{Plummer2003JAGS}. These tools enable non-specialist data scientists to perform Bayesian modelling and computation in a more efficient manner, specifically in terms of development time and programming effort.
Stan is a probabilistic programming language that is focused on Bayesian statistical modelling. The underlying computational algorithm used by Stan to sample from the posterior distribution is a variation of the Hamiltonian
MCMC algorithm \citep{Neal1994improved}, known as the ``No U-Turn Sampler'' (NUTS). This procedure uses autodifferentiation to automatically build an MCMC sampler from the posterior distribution specified by the user. Conveniently, Stan is also integrated into R through the package \emph{Rstan} \citep{SDT2018RStan}.

Although Stan offers a number of advantages as a state-of-the-art Bayesian tool, the
primary benefit of the tool stems from the separation between modelling (through the probabilistic programming language) and computational
processes \citep{Betancourt2017conceptual} (through the built-in Hamiltonian sampling algorithm) that it provides. This separation enables users to focus primarily on the
specification of the problem, i.e., the choice of model and prior distributions, rather than the computational aspects of sampling. For most problems, users are not required to specify any computational procedures,
as they can be handled adequately by the default process. However, the algorithm is not fool-proof, and in general both run-time and convergence of the resulting Hamiltonian sampler can be heavily influenced by the way in which models and priors are specified.

In addition to Stan, there are a number of other probabilistic programming tools that support Bayesian automated MCMC sampling. In particular, JAGS (and its predecessor BUGS) has also been extensively used by Bayesian data analysts \citep{Lunn2000WinBUGS}. In comparison to Stan, JAGS uses an automatic Gibbs sampling approach which can be less efficient than the NUTS algorithm used by Stan~\citep{R-bloggers2014JAGS}.

\section{Local and Global Trend non-seasonal model}\label{sec:non-seas}

To address the shortcomings identified in Section~\ref{sec:short}, we propose the non-seasonal local and global trend model (LGT, note that ``local'' and ``global'' refers here to parts of the series or the whole series, as opposed to models that learn across time series and are also oftentimes called global models, see, e.g., \cite{januschowski2020criteria}). 
It is derived from the classical ETS model with additive error, additive trend, and no seasonality, with the addition of a number of distinguishing features. We define it as follows.
\begin{align}
\label{equ:lgt1} y_{t+1} &\sim t(\nu,\hat{y}_{t+1}, \hat{\sigma} _{t+1}),\\
\label{equ:lgt2} \hat{y}_{t+1} &= l_{t}+ \gamma l_{t}^{ \rho }+ \lambda b_{t},\\
\label{equ:lgt3} l_{t+1} &= \alpha y_{t+1}+ \left( 1- \alpha  \right) l_{t},\\
\label{equ:lgt4} b_{t+1} &= \beta  \left( l_{t+1}-l_{t} \right) + \left( 1- \beta  \right) b_{t},\\
\label{equ:lgt5} \hat{\sigma}_{t+1} &= \sigma \hat{y}_{t+1}^{ \tau}+ \xi.
\end{align}
Here, $t(v,m,s)$ denotes a Student $t$-distribution with degrees-of-freedom $v$, location $m$ and scale $s$. Furthermore, $y_{t}$ denotes the value of the dependent variable of interest at time $t$, $\hat{y}_{t+1}$ is the predicted value (conditional mode) of $y$ at time $t+1$ given information up to time $t$; $\hat{\sigma}_{t+1}$ denotes the scale of errors at time $t+1$; $l_{t}$ is the level at time $t$; and $b_{t}$ the local trend at time $t$. The parameters that must be estimated are described in Table~\ref{tab:lgtParam}.

\begin{table}[!htb]
\centering
\begin{tabular}{ll}
\hline
         & Description \\ \hline
$\nu$ & degrees of freedom of the $t$-distribution\\
$\gamma$ & coefficient of the global trend \\
$\rho$ & power coefficient of the global trend,\\
       & by default in $[-0.5, 1]$ \\ 
$\lambda$ & damping coefficient of the local trend, in $[0, 1]$ \\
$\alpha$ & smoothing parameter for the level term, \\ & in $[0, 1]$ \\
$\beta$ & smoothing parameter for the local trend \\ & term, in $[0, 1]$ \\
$\sigma$ & coefficient of the size of error, positive \\
$\tau$ & power coefficient of the size of error, in $[0, 1]$\\
$\xi$ & minimum value of the size of error, positive \\
$b_1$ & initial local trend \\
\hline
\end{tabular}
\caption{Parameters to be fitted for the LGT model}
\label{tab:lgtParam}\end{table}

Equation~\ref{equ:lgt1} models observation $y_{t+1}$ by a Student $t$-distribution, where Equation~\ref{equ:lgt2} defines the conditional mode (i.e., the predicted value) of the Student $t$-distribution at the next time-step. Equation~\ref{equ:lgt3} estimates the level adjustment via exponential smoothing, and Equation~\ref{equ:lgt4} estimates the local trend via exponential smoothing of the first differences of the smoothed levels, and Equation~\ref{equ:lgt5} defines the level dependent (heteroscedastic) scale of error distribution.

Equation~\ref{equ:lgt3} is very reminiscent of the Holt linear method with a slightly simpler structure. 
The local trend update formula is the same as in the Holt linear (AAN) method, see Equation~\ref{equ:aan4}. The main differences are now discussed in more detail.

\subsection{Student $t$-distributed errors}

Equation~\ref{equ:lgt1} models the observations via a Student $t$-distribution, with a time-varying scale. We note that the Student $t$-distribution has three free parameters as opposed to the two free parameters of the normal distribution. The additional parameter, $\nu$, commonly referred to as the degrees-of-freedom, controls the weight of the tails of the distribution, which facilitates the graceful handling of ``outliers'' in data.
This choice of data distribution is in contrast with the innovation state space models for  classical ETS \citep{Hyndman2018Forecasting}, in which the error distribution is assumed to be normal and identically distributed with a mean of zero and variance of $\sigma^2$.

However, as outlined in Section~\ref{sec:short}, this assumption is often overly restrictive in practice. This motivates our choice to generalise the error assumption from normal to Student $t$.
The Student $t$-distribution is a generalisation of the normal distribution, with the heaviness of the tails being smoothly controlled by the degrees-of-freedom parameter $\nu$. When $\nu$ is relatively high (i.e., above 20), the Student $t$ is virtually identical to a normal distribution. In contrast, when $\nu$ is small the tails of the $t$-distribution are heavier than the tails of a normal distribution; when $\nu=1$ the $t$-distribution reduces to the special case of the Cauchy distribution.

Like all other model parameters, $\nu$ is fitted automatically within our Bayesian approach. This means that the model is able to automatically adapt to relatively smooth time series without large spikes (e.g., like those of the M3 competition), but can handle the more volatile series that are often found in practical applications. 

We also note that in principle our models can deal with asymmetric distributions. However, the results of preliminary experiments indicated that the fitting was slower and results were not improved, so we did not pursue this option further.

\subsection{One-step ahead prediction}  

In the classical AAN model (Equation~\ref{equ:aan2}), the forecasts are calculated as $\hat{y}_{t+1} = l_{t} + b_{t}$, where $l_{t}$ is the current estimate of the level and $b_{t}$ is the current estimate of the trend, resulting in a linearly changing forecast. In multiplicative models the one-step-ahead prediction is instead given by $\hat{y}_{t+1} = l_{t} \, b_{t}$ \citep{Hyndman2018Forecasting}, which results in an exponentially varying forecast.
In the latter case, for a model to be reasonably stable, the trend values $b_t$ need to be close to one. Define the trend to be $b_t=1 + \delta_t$, where $\delta_t$ is a small value; then $\hat{y}_{t+1} = l_{t} b_{t} = l_{t} \left(1 + \delta_{t}\right) = l_{t} + l_{t} \delta_{t}$. This suggests that both the additive and multiplicative case can be generalised by the following equation:
\begin{align}
\label{equ:gen1}
\hat{y}_{t+1}=l_{t} + b_{t} l_{t}^{\rho},
\end{align}
with $\rho \in \left[0, 1\right]$. While this generalisation of additive and multiplicative trend model is interesting, we found in practice that its performance was typically worse than that of the alternative global trend model:
\begin{align}
	\label{equ:gen2}
\hat{y}_{t+1}=l_{t} + \gamma l_{t}^{\rho},
\end{align}
where $\gamma$ is constant for the whole time series. 
This flexible form of trend function is particularly useful for dealing with time series that grow faster than linear but slower than exponential. If the $\rho$ parameter is allowed to be negative, then for a growing trend we can obtain a damped behaviour. In practice, e.g., when fitting to the M3 time series, we did not observe this capability to be particularly important.

Additionally, the model includes a local linear trend of the form $\lambda b_{t}$, where $\lambda$ is a value between $-1$ and $1$. In practice, this trend is almost always between $0$ and $1$. Restricting the magnitude of this coefficient means that the model is sensitive to the recent changes in slope, but to a lesser extent than the original AAN model. A negative $\lambda$ coefficient is indicative of a counter-trending behaviour, but in practice this appears to be a relatively rare situation.
The one-step-ahead forecast is then: 
\begin{align}
	\label{equ:gen3}
\hat{y}_{t+1} = l_{t} + \gamma l_{t}^{\rho} + \lambda b_{t}.
\end{align}
It is assumed, and enforced during the simulation (i.e., forecasting) stage, that the level and forecasting values are all strictly positive, and larger than 0.001 to avoid numerical instabilities. All model parameters, such as $\gamma$, $\rho$, and $\lambda$, are fitted via MCMC sampling from the posterior distribution.

The one-step-ahead forecast given by Equation~\ref{equ:gen3}, is given by the linear combination of the level $l_{t}$, the global nonlinear trend $\gamma l_{t}^{ \rho }$, and the damped local linear trend $\lambda b_{t}$. We use the term ``global'' to describe the nonlinear trend to stress the fact that both $\gamma$ and $\rho$ are fitted on the entire time series and do not change over time. 
The simple function that defines the global trend has several interesting properties. If $\rho$ is close to zero, the contribution of the global trend to the one-step-ahead prediction becomes near-constant (a drift term), and the resulting trend becomes close to linear, leading to an overall (global) upward or downward trend.
If $\rho$ is not zero, the global trend is a time-invariant function of the level. The level is typically smooth but puts more weight on more recent observations. As such one could argue that there is a local component in the ``global'' trend. However, the global trend would still be smooth and changing slowly, whereas the local trend component can change much quicker. Typically, the global trend will dominate eventually over the short-term behaviour caused by the local linear trend. This combination of local and global trend can be especially useful, for example, in situations like the Covid-19 pandemic, as the model is able in such a situation to model both long-term overall trend and short-term changes in the behaviour of the series.
Finally, if $\rho$ is close to one, the contribution of the global trend to the one-step-ahead prediction becomes almost proportional to the level value, and the trajectory of the multistep forecast becomes exponential. This means that the global trend is flexible enough to adequately model trends typical of fast growing businesses, i.e., growth that is faster than linear but slower than exponential in nature.

\subsection{Heteroscedastic error}

In addition to accounting for a possible heavy-tailed error distribution, our proposed model also allows the scale of the error to vary as a function of the estimated level of the time series.
This is achieved by allowing the scale parameter of the Student's $t$-distribution to vary according to a monotonically increasing function of the current level of the time series. This is given by Equation~\ref{equ:lgt5}, which models the scale of the error by a power transformation of the one-step-ahead prediction plus a level-independent base scale term $\xi$. 

This allows us to account for the common situation in which the magnitude of the error is an increasing function of value of the time series, with the use of a variable power $\tau$ allowing us to moderate the rate at which the scale increases as the value of the time series increases. In practice, the value of $\tau$ is usually limited to values between zero and one. A $\tau$ of zero corresponds to the situation of a constant scale across the series, i.e., homoscedasticity. In contrast, $\tau=1$ approximates the behaviour of the multiplicative error ETS model, in which the magnitude of the error is proportional to the expected value. In practice, a value between zero and one is desirable for $\tau$ as we expect the error to grow in relation to the magnitude of the time series, but in a sub-linear manner. The parameter $\xi>0$ sets the base error scale. These modifications relax the usual NID assumption, as the error term is now not required to be either normally or identically distributed.

\section{Seasonal Global Trend Model}\label{sec:seas}

We also consider the extension of our procedure to seasonal models. Specifically, we introduce the seasonal global trend model (SGT), which is a seasonal modification of the LGT model previously discussed. In contrast to the non-seasonal model, we remove the local linear trend term and replace it with the ability to model seasonal effects. 
Though the local linear trend could be maintained, in preliminary experiments not reported here we did not find a benefit from having the local trend, and fitting such models was slower. Our hypothesis is that the models get too complex with too many parameters then.
We utilise a multiplicative seasonality scheme, with the seasonality coefficients evolving as per the usual ETS model:
\begin{eqnarray}
l_{t} &=& \alpha \cdot \frac{y_t}{s_t} + (1-\alpha)\cdot l_{t-1} \\
s_{t+m} &=& \zeta \cdot \frac{y_t}{l_t} + (1-\zeta)\cdot s_{t}
\end{eqnarray}
Here, $m>0$ is the seasonality (e.g., $m=12$ for a monthly series), $\zeta \in [0,1]$ is the seasonality smoothing coefficient, and $s_t$ are seasonality factors.

A key difference in our Bayesian implementation, {\em vis \`{a} vis} the conventional seasonal ETS model, is the manner in which the initial seasonality coefficients, $s_1,\ldots,s_m$, are handled. Rather than calculate the ratio of observations against a cyclical mean for a few cycles, potentially followed by maximum likelihood estimation, they are handled in a fully Bayesian fashion. That is, we simply treat these quantities as unknown parameters and assign them appropriate prior distributions; specifically $\log s_i \sim C(0,b)$, where $C(a,b)$ denotes a Cauchy distribution with location $a$ and scale $b$. We then allow these quantities to be sampled along with the other model parameters as part of the MCMC sampling procedure. 

The complete specification of the  model is:
\begin{align}
	\label{equ:sgt1} y_{t+1} &\sim t (\nu,\hat{y}_{t+1}, \hat{\sigma} _{t+1})\\
	\label{equ:sgt2} \hat{y}_{t+1} &= (l_{t}+ \gamma l_{t}^{ \rho })s_{t+1}\\
	\label{equ:sgt3} l_{t+1} &= \alpha \frac{y_{t+1}}{s_{t+1}} + \left( 1- \alpha  \right) l_{t}\\
	\label{equ:sgt4} s_{t+m} &= \zeta \cdot \frac{y_t}{l_t} + (1-\zeta)\cdot s_{t}\\
	\label{equ:sgt5} \hat{\sigma}_{t+1} &= \sigma \hat{y}_{t+1}^{ \tau}+ \xi
\end{align}
with
\[
    \frac{1}{m} \sum_{i=1}^m s_i = 1.
\]
The latter condition enforces the requirement that the seasonality adjustments be expressed relative to the overall scale of the signal. 
The complete set of parameters that need to be fitted for the SGT model are described in Table~\ref{tab:sgtParam}.

\begin{table}[!htb]
	\centering
	\begin{tabular}{ll}
		\hline
		& Description \\ \hline
		
		$\nu$ & degrees of freedom of the t-distribution\\
		$\gamma$ & coefficient of the global trend \\
		$\rho$ & power coefficient of the global trend,\\
		& by default in $[-0.5, 1]$ \\ 
		
		$\alpha$ & smoothing parameter for the level term, \\ & in $[0, 1]$ \\
        $\zeta$ & smoothing parameter for the seasonality terms \\
        & in $[0,1]$ \\
		$\sigma$ & coefficient of the size of error, positive \\
		$\tau$ & power coefficient of the size of error, in $[0, 1]$\\
		
		$\xi$ & minimum value of the size of error, positive \\
		
		$s_i$ & initial seasonality, positive; for $i$ in $\{1, \dots, m\}$ \\
		\hline
	\end{tabular}
	\caption{Parameters to be fitted for the SGT model}
	\label{tab:sgtParam}\end{table}

During the forecasting (simulation) phase, the seasonal coefficients are not updated, using a frozen setup of the last $m$ factors. The motivation for this process is that these coefficients are meant to adjust to changing situations (new data), but during the forecasting phase we do not have new data available.

\section{Implementation}

In Sections 5 and 6 we introduced several extensions to the classical ETS model, that allow for a wider range of trend and error models. We fit these models within the Bayesian framework of statistical inference. Specifically, we use MCMC approaches to simulate from the posterior distribution, and use these to approximate the posterior predictive distribution over forecasts. We provide two implementations: (i) a Stan-based implementation, and (ii) a bespoke Gibbs sampling implementation based on a simplified model. The former is more flexible and easier for users to extend, but is quite slow. The Gibbs sampler is less straightforward to extend, but is substantially faster and does not require a working Stan installation to run. In the following we detail the approaches.

\subsection{Prior Distributions and Parameter Space}

A key step in Bayesian inference is the specification of prior distributions, from which the model parameters are randomly sampled. The choice of prior distributions can substantially affect the performance of a Bayesian procedure, particularly for small sample sizes, and a number of these prior distributions are parametrized by their own parameters, so a user can influence the initialisation and fitting. In our case, the random sampling from the prior distributions replaces the relatively sophisticated initial choices for ETS model parameters in traditional ETS models. 

Also the bounds within which the parameters can be set have received considerable attention in the literature for ETS models \citep{Hyndman2018Forecasting,Hyndman2008Forecasting,hyndman2008admissible}. The parameters are traditionally restricted to the intervals $0<\alpha<1$, $0<\beta<\alpha$, and $0<\zeta<1-\alpha$ (the seasonal parameter is called $\gamma$ in that literature, but we use $\zeta$ in our paper). The damping parameter $\phi$ is constrained, e.g., to $0.8<\phi<0.98$, to prevent numerical problems \citep{Hyndman2018Forecasting}.
Furthermore, there are bounds for forecastability and stability conditions. For the additive ETS model variants on which we focus, these are less tight than the traditional restrictions \citep{Hyndman2018Forecasting,hyndman2008admissible}, so not relevant for our case. Finally, for the seasonal models, ETS (in its implementation in the \emph{forecast} package) also checks that the roots of the characteristic equation all lie within the unit circle (also see \cite{hyndman2008admissible}).

As our models have several modifications in comparison with the traditional ETS models, other parameter region constraints may be applicable. However, we do not perform a detailed theoretical study of these bounds, and instead choose bounds that are in line with traditional ETS and that empirically lead to good forecasts. In particular, we follow \cite{Bermudez2010Bayesian}, who have presented a Bayesian exponential smoothing approach for an additive Holt-Winters (AAA) model. Those authors constrain the smoothing parameters to lie in the interval (0,1), and use non-informative priors.

We use the following priors in our implementation. 
The degrees-of-freedom $\nu$ and power-coefficients $\rho$ and $\tau$ are assigned uniform distributions over appropriate intervals; the scale $\sigma$ and minimum error scale $\xi$ are assigned half-Cauchy distributions; the smoothing parameters $\alpha$, $\beta$ and $\zeta$ (for seasonal models) are assigned beta distributions; the coefficients $\gamma$ and $\lambda$ are assigned heavy-tailed, weakly informative Cauchy distributions~\citep{Gelman2006}; and the initial values for the seasonality terms, $s_1,\ldots,s_m$ are assigned log-Cauchy prior distributions. The hyperparameters for these prior distributions are given default values in our implementation and there is in principle no need for the user to specify these. 

However, the choice of priors can potentially greatly influence the model outcomes, and the user might have prior knowledge to be incorporated in the modelling in the form of a tightened prior. Due to the flexibility of Stan it is straightforward for the user to modify these prior hyperparameter values if so desired.

\subsection{Gibbs Sampler}

In addition to providing a Stan implementation, we also developed an MCMC sampling algorithm for our proposed model via a bespoke Gibbs sampler. 

\subsubsection{Modified Model}

For purposes of tractability, the Gibbs sampler is based on a slightly simplified form of the LGT model:
\begin{eqnarray*}
    y_{t+1} \vbar \hat{y}_{t+1}, 
    \sigma, \nu &\sim& t(\nu, 
    \hat{y}_{t+1}, 
    \sigma), \\
    \hat{y}_{t+1} &=& l_t + \gamma l_t^\rho + \lambda b_t, \\
    l_{t} &=& \alpha y_t + (1-\alpha) l_{t-1}, \\
    b_{t} &=& \beta (l_t - l_{t-1}) + (1-\beta) b_{t-1}.
\end{eqnarray*}
The simplified model enforces homoscedasticity on the errors, which dramatically improves the convergence speed of the resulting Gibbs sampler.
For the purposes of sampling, the Gibbs sampler represents the $t$-distribution via a scale-mixture of normal distributions~\citep{LangeEtAl89}, i.e., 
\begin{eqnarray*}
    y_{t+1} \vbar \mu_{t}, \sigma, \omega_{t+1}^2  &\sim& N(\mu_{t}, \sigma^2 \omega_{t+1}^2), \\    \omega_{t+1}^2 \vbar \nu &\sim& {\rm IG}\left( \frac{\nu}{2}, \frac{\nu}{2} \right),
\end{eqnarray*}
where ${\rm IG}(a,b)$ denotes the inverse-gamma distribution.
This representation is equivalent, as integrating out the latent variables $\omega_t^2$ exactly recovers the $t$-distribution. Despite introducing $n$ new latent variables that need to be sampled, this representation lets us treat $\hat{y}_t$ as the linear predictor in a regular Gaussian linear regression, making sampling $\gamma$ and $\lambda$ substantially easier. We also utilised this same scale-mixture form to represent the Cauchy prior distributions, as well as an inverse-gamma scale mixture of inverse-gammas to represent the half-Cauchy prior distribution~\citep{WandEtAl11}.

A second difference between the Gibbs sampler and the Stan implementation is in the way in which the $\rho$ and $\nu$ parameters are handled. Given the non-standard form of the conditional distributions for these parameters, the Gibbs sampler utilises a simple grid-sampler by restricting the range of possible values these parameters can take to a finite, discrete set. We found this has little effect on the performance of the methods, as the models appear quite insensitive to the specific values of these parameters, but substantially improves convergence of the resulting algorithm. 

\subsubsection{The Sampling Algorithm}

We now briefly summarise the Gibbs sampling algorithm. Gibbs sampling from a posterior distribution, say $p(\theta_1,\theta_2 \vbar y)$ works by iteratively sampling from the conditional distributions $\theta_1 \vbar \theta_2, y$ and $\theta_2 \vbar \theta_1, y$. This can be efficient when it is relatively easy to draw random samples from the conditional distributions. Let $\hat{y} = (\hat{y}_1,\ldots,\hat{y}_n)$ denote the vector of one-step-ahead forecasts made by the LGT model, and $\omega = (\omega_1,\ldots,\omega_n)$ denote the vector of latent variables. We now summarise the specific Gibbs sampling steps:
\begin{enumerate}
    \item Sample $\sigma^2 \vbar y, \hat{y}, \omega$. Due to the use of the scale-mixture representation for the $t$-distribution the half-Cauchy prior, this conditional distribution is of the inverse-gamma type.
    
    \item Sample the latent variables ${\omega} \vbar y, \hat{y}, \nu, \sigma^2$; the conditional distributions for these are also of the inverse-gamma type~\citep{LangeEtAl89}.
    
    \item Sample $\nu \vbar \omega$ using a grid-sampler. 
    
    \item Sample the coefficients $\gamma,
    \lambda \vbar y, \hat{y}, \omega, \sigma^2$; due to the latent-variable representation of the likelihood and prior distributions,  the conditional distributions for these are just (truncated) normal distributions.
    
    \item Sample $\alpha, \beta \vbar y, \hat{y}, \nu, \sigma^2$ using an adaptive Metropolis-Hastings sampler~\citep{SchmidtMakalic20a}. For this step we integrate the latent variables $\omega$ out of the likelihood to improve convergence of the Gibbs sampler.
    
    \item Sample the power coefficient $\rho \vbar \hat{y}, \nu, \sigma^2$ using a grid-sampler, again integrating out the latent variables $\omega$. 
\end{enumerate}
These steps are iterated repeatedly until the desired number of samples have been generated.

\section{Experiments}\label{sec:performance}

As our proposed models are extensions of the traditional exponential smoothing model family, they are per-series univariate models in nature and do not train across series, or utilise additional covariates. As such, we opt for an evaluation on the classical M3 benchmark dataset. This has been the standard benchmark of datasets in forecasting for decades, and remains an exellent and highly representative dataset for evaluating univariate methods of the type presented in this paper. Thus, the M3 dataset is the focus of our experiments.

 We use the M3 time series and forecasts from the ``\emph{Mcomp}'' R package \citep{Hyndman2013Mcomp}.
The M3 dataset consists of a total of 3003 time series in the four categories yearly, quarterly, monthly, and other. Details of the dataset are given in Table~\ref{tab:M3}. The quarterly and monthly categories are seasonal at different levels of seasonality, and the remaining two categories are non-seasonal. 

In the absence of a more sophisticated model selection mechanism, as, e.g., ETS has it (and that could further improve the results of our model), for the two categories that are assumed to be seasonal, we use the SGT model, whereas for the non-seasonal categories we use the LGT model.

\begin{table}[!tb]
\centering
\begin{tabular}{lrrrr}
\hline
$f$ & \#$s$ & $l_{min}$ & $l_{max}$ & $h$\\
\hline
Yearly & 645 & 20 & 47 & 6\\
Quarterly & 756 & 24 & 72 & 8\\
Monthly & 1428 & 66 & 144 & 18\\
Other & 174 & 71 & 104 & 8\\
\hline
Total & 3003 &&&\\
\hline
\end{tabular}
\caption{Dataset characteristics of the M3 dataset. Here, $f$ is the frequency, \#$s$ the amount of time series, $l_{min}$ and $l_{max}$ are the minimum and maximum length of the series, and $h$ is the forecast horizon.} 
\label{tab:M3}
\end{table}

\subsection{Error measures}

When evaluating our new algorithms in terms of point forecasting, we use the following two error measures: (i) the symmetric mean absolute percentage error (sMAPE), which was the primary measure used in the M3 competition \citep{Makridakis2000M3}, given by Equation~\ref{equ:smape}, and (ii) the mean absolute scaled error \citep[MASE, ][]{Hyndman2006Another}, given by Equation~\ref{equ:mase}.
\begin{equation}
\label{equ:smape}
\text{sMAPE} = \frac{200}{h} \sum_{t=1}^h \frac{\left|y_{n+t} - \hat{y}_{n+t}\right|}{\left| y_{n+t}\right| + \left| \hat{y}_{n+t}\right|},
\end{equation}
\begin{equation}
\label{equ:mase}
\text{MASE}=\frac{h^{-1}\sum_{t=1}^h \left|y_{n+t} - \hat{y}_{n+t}\right|}
{(n-s)^{-1}\sum_{t=s+1}^n \left|y_t - y_{t-s}\right|}.
\end{equation}
Here, $h$ is the maximum prediction horizon, $s$ is the seasonality ($s=1$ for non-seasonal series and $s>1$ for seasonal series, e.g., $s=4$ for quarterly series), and $n$ is the length of the time series.

The motivation to use these two measures is as follows. We use the sMAPE as it is the error measure used in the original M3 competition, to ensure that the results are directly comparable, and to avoid any irregularities that have been reported in the literature \citep{Boylan2015Reproduction}. Even though sMAPE has well-documented problems (see, e.g., \cite{Hyndman2006Another}), we argue that it may be the fairest metric to compare with the original participants, as they would have optimised their methods towards this metric. The MASE is used as a measure that is accepted nowadays in the forecasting community as a measure that does not have such problems.

For probabilistic forecasting, we use the Mean Scaled Interval Score (MSIS) as defined in Equation~\ref{equ:msis}, which is designed to evaluate prediction intervals. The interval score was first proposed by \cite{gneiting2007strictly}, and computes the sum over the size of the intervals and the magnitude of error for points that lie outside of the interval. The MSIS as used by \cite{Makridakis2018M4}, which we follow here, then achieves a scale-free measure by dividing by the in-sample na\"ive forecast in the same way as the MASE.
In Equation~\ref{equ:msis}, $q_t^{[u]}$ and $q_t^{[l]}$ are the upper and lower bounds of the prediction interval, respectively, $\mathbbm{1}$ is the indicator function that has a value of one if the condition is true, zero otherwise, and $\alpha$ is the significance level, i.e., it defines the target width of the interval. For example, for 90\% prediction intervals, we set $\alpha$ to 0.1.

\begin{figure*}[t]
\begin{equation}
\label{equ:msis}
\text{MSIS} = \frac{
h^{-1}\sum_{t=n+1}^{n+h} \left( {q_t^{[u]}-q_t^{[l]}}+\frac{2}{\alpha}(q_t^{[l]}-y_t)\mathbbm{1}_{y_t<q_t^{[l]}}+\frac{2}{\alpha}(y_t-q_t^{[u]})\mathbbm{1}_{y_t>q_t^{[u]}} \right)
}{
(n-s)^{-1}\sum_{t=s+1}^{n}{|y_t-y_{t-s}|}
}
\end{equation}
\end{figure*}

\subsection{Statistical testing}

We use statistical testing and critical difference diagrams as proposed by \cite{demsar2006statistical} and later extended by \cite{benavoli2016should}, in a Python implementation of \cite{Bunse2023critdd}. In particular, we use a Friedman test to find differences in the outcomes and then a Wilcoxon signed-rank test with Holm adjustment for multiple comparisons, to characterise the differences. The statistical significance results are visualised in a critical difference diagram where a black line connecting the methods means that no statistical significance between their results could be determined.

\subsection{Results and comparison}

To facilitate a fair assessment of our proposed univariate methods we restrict the scope of competitor methods to per-series univariate techniques only, excluding from our comparison any methods that utilise across-series information. This is consistent with an application case when there is only a single time series or a small amount of series available to be forecasted and thus no across-series modelling is possible.

\begin{table*}[!tb]
\centering\footnotesize
\begin{tabular}{lrr}
\hline
\hline
~
 &
sMAPE &
MASE \\\hline
\multicolumn{3}{c}{\centering \textit{All (amount of series: 3003)}}\\\hline
\{L,S\}GT & \textbf{12.29}, sd=0.010 & \textbf{1.30}, sd=0.0007\\
Best algorithm in M3 & 12.76 (THETA) & 1.39 (THETA) \\
Hybrid & 12.82 & 1.40 \\
ETS/ZZZ & 13.07 & 1.43 \\
ARIMA & 13.58 & 1.46 \\
MAPA & 12.71 & 1.41 \\
THETA & 12.79 & 1.42\\
baggedETS & 12.95 & 1.42\\\hline

\multicolumn{3}{c}{\centering \textit{Yearly (amount of series: 645, horizon: 6)}}\\\hline
LGT & \textbf{15.18}, sd= 0.015 & \textbf{2.48}, sd= 0.004 \\
Best algorithm in M3 & 16.42 \ (RBF) & 2.63  (ROBUST-Trend) \\
Best of baggedETS (orig) & 17.80 (BLD.Sieve) & 3.15 (BLD.MBB) \\
Hybrid & 16.73 & 2.85 \\
ETS/ZZZ & 17.00 & 2.86 \\
ARIMA & 17.12 & 2.96 \\
THETA & 16.75 & 2.77 \\
MAPA & 16.91 & 2.86 \\
baggedETS & 17.18 & 2.88\\\hline

\multicolumn{3}{c}{\centering \textit{Other (amount of series: 174, horizon: 8)}}
\\\hline
LGT & \textbf{4.25}, sd=0.017 & \textbf{1.72}, sd=0.002 \\
Best algorithm in M3 & 4.38 (ARARMA) & 1.86 (AutoBox2) \\
Hybrid & 4.33 & 1.79 \\
ETS/ZZZ & 4.37 & 1.81 \\
ARIMA & 4.46 & 1.83 \\
THETA & 4.92 & 2.27 \\
MAPA & 4.48 & 1.87 \\
baggedETS & 4.72 & 1.96\\\hline

\multicolumn{3}{c}{\centering \textit{Monthly (amount of series: 1428, horizon: 18)}}
\\\hline
SGT & 13.77, sd=0.019 & \textbf{0.83}, sd=0.0006 \\
Best algorithm in M3 & 13.89 (THETA) & 0.85 (ForecastPro) \\
Best of baggedETS (orig) & 13.64 (BLD.MBB) & 0.85 (BLD.MBB) \\
Hybrid & 13.91 & 0.84 \\
ETS/ZZZ & 14.14 & 0.86 \\
ARIMA & 14.98 & 0.88 \\
THETA & 13.86 & 0.86 \\
MAPA & \textbf{13.62} & 0.84 \\
baggedETS & 13.71 & \textbf{0.83}\\\hline

\multicolumn{3}{c}{\centering \textit{Quarterly (amount of series: 756, horizon: 8)}}
\\\hline
SGT & \textbf{8.87}, sd=0.011 & \textbf{1.07}, sd=0.0019 \\
Best algorithm in M3 & 8.96 (THETA) & 1.09 (THETA) \\
Best of baggedETS (orig) & 10.03 (BLD.Sieve) & 1.22 (BLD.MBB) \\
Hybrid & 9.39 & 1.14 \\
ETS/ZZZ & 9.68 & 1.17 \\
ARIMA & 10.00 & 1.19 \\
THETA & 9.20 & 1.11 \\
MAPA & 9.31 & 1.13 \\
baggedETS & 9.81 & 1.16\\\hline

\end{tabular}
\caption{Overall point prediction performance on the M3 dataset.} 
\label{tab:overall}
\end{table*}

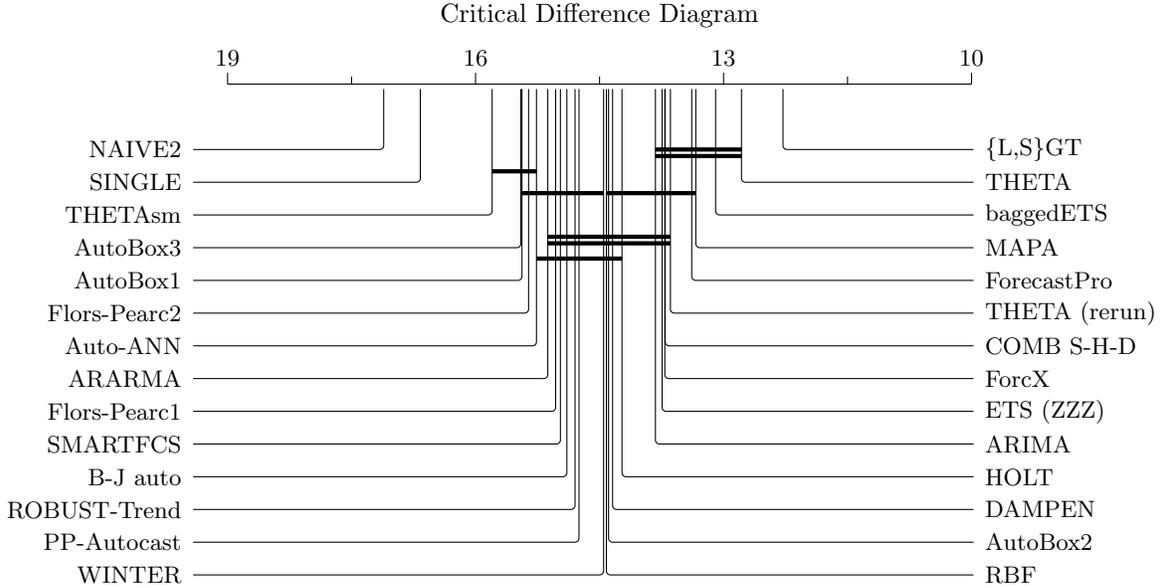
\begin{figure*}[htb]
\begin{tikzpicture}[
  treatment line/.style={rounded corners=1.5pt, line cap=round, shorten >=1pt},
  treatment label/.style={font=\small},
  group line/.style={ultra thick},
]

\begin{axis}[
  clip={false},
  axis x line={center},
  axis y line={none},
  axis line style={-},
  xmin={10},
  ymax={0},
  scale only axis={true},
  width={0.6\textwidth},
  ticklabel style={anchor=south, yshift=1.3*\pgfkeysvalueof{/pgfplots/major tick length}, font=\small},
  every tick/.style={draw=black},
  major tick style={yshift=.5*\pgfkeysvalueof{/pgfplots/major tick length}},
  minor tick style={yshift=.5*\pgfkeysvalueof{/pgfplots/minor tick length}},
  title style={yshift=\baselineskip},
  xmax={19},
  ymin={-15.5},
  height={16\baselineskip},
  xtick={1,4,7,10,13,16,19,22,25,28},
  minor x tick num={1},
  x dir={reverse},
  title={Critical Difference Diagram},
]

\draw[treatment line] ([yshift=-2pt] axis cs:12.28071928071928, 0) |- (axis cs:9.947385947385946, -2.0)
  node[treatment label, anchor=west] {\{L,S\}GT};
\draw[treatment line] ([yshift=-2pt] axis cs:12.783050283050283, 0) |- (axis cs:9.947385947385946, -3.0)
  node[treatment label, anchor=west] {THETA};
\draw[treatment line] ([yshift=-2pt] axis cs:13.096070596070597, 0) |- (axis cs:9.947385947385946, -4.0)
  node[treatment label, anchor=west] {baggedETS};
\draw[treatment line] ([yshift=-2pt] axis cs:13.335497835497835, 0) |- (axis cs:9.947385947385946, -5.0)
  node[treatment label, anchor=west] {MAPA};
\draw[treatment line] ([yshift=-2pt] axis cs:13.384781884781885, 0) |- (axis cs:9.947385947385946, -6.0)
  node[treatment label, anchor=west] {ForecastPro};
\draw[treatment line] ([yshift=-2pt] axis cs:13.642690642690642, 0) |- (axis cs:9.947385947385946, -7.0)
  node[treatment label, anchor=west] {THETA (rerun)};
\draw[treatment line] ([yshift=-2pt] axis cs:13.705960705960706, 0) |- (axis cs:9.947385947385946, -8.0)
  node[treatment label, anchor=west] {COMB S-H-D};
\draw[treatment line] ([yshift=-2pt] axis cs:13.710456210456211, 0) |- (axis cs:9.947385947385946, -9.0)
  node[treatment label, anchor=west] {ForcX};
\draw[treatment line] ([yshift=-2pt] axis cs:13.743090243090244, 0) |- (axis cs:9.947385947385946, -10.0)
  node[treatment label, anchor=west] {ETS (ZZZ)};
\draw[treatment line] ([yshift=-2pt] axis cs:13.824009324009324, 0) |- (axis cs:9.947385947385946, -11.0)
  node[treatment label, anchor=west] {ARIMA};
\draw[treatment line] ([yshift=-2pt] axis cs:14.22860472860473, 0) |- (axis cs:9.947385947385946, -12.0)
  node[treatment label, anchor=west] {HOLT};
\draw[treatment line] ([yshift=-2pt] axis cs:14.342324342324343, 0) |- (axis cs:9.947385947385946, -13.0)
  node[treatment label, anchor=west] {DAMPEN};
\draw[treatment line] ([yshift=-2pt] axis cs:14.391275391275391, 0) |- (axis cs:9.947385947385946, -14.0)
  node[treatment label, anchor=west] {AutoBox2};
\draw[treatment line] ([yshift=-2pt] axis cs:14.417249417249417, 0) |- (axis cs:9.947385947385946, -15.0)
  node[treatment label, anchor=west] {RBF};
\draw[treatment line] ([yshift=-2pt] axis cs:14.451048951048952, 0) |- (axis cs:19.444555444555444, -15.0)
  node[treatment label, anchor=east] {WINTER};
\draw[treatment line] ([yshift=-2pt] axis cs:14.748584748584749, 0) |- (axis cs:19.444555444555444, -14.0)
  node[treatment label, anchor=east] {PP-Autocast};
\draw[treatment line] ([yshift=-2pt] axis cs:14.796536796536797, 0) |- (axis cs:19.444555444555444, -13.0)
  node[treatment label, anchor=east] {ROBUST-Trend};
\draw[treatment line] ([yshift=-2pt] axis cs:14.895271395271395, 0) |- (axis cs:19.444555444555444, -12.0)
  node[treatment label, anchor=east] {B-J auto};
\draw[treatment line] ([yshift=-2pt] axis cs:14.973526473526473, 0) |- (axis cs:19.444555444555444, -11.0)
  node[treatment label, anchor=east] {SMARTFCS};
\draw[treatment line] ([yshift=-2pt] axis cs:15.031302031302031, 0) |- (axis cs:19.444555444555444, -10.0)
  node[treatment label, anchor=east] {Flors-Pearc1};
\draw[treatment line] ([yshift=-2pt] axis cs:15.126207126207126, 0) |- (axis cs:19.444555444555444, -9.0)
  node[treatment label, anchor=east] {ARARMA};
\draw[treatment line] ([yshift=-2pt] axis cs:15.263236763236764, 0) |- (axis cs:19.444555444555444, -8.0)
  node[treatment label, anchor=east] {Auto-ANN};
\draw[treatment line] ([yshift=-2pt] axis cs:15.358474858474858, 0) |- (axis cs:19.444555444555444, -7.0)
  node[treatment label, anchor=east] {Flors-Pearc2};
\draw[treatment line] ([yshift=-2pt] axis cs:15.43839493839494, 0) |- (axis cs:19.444555444555444, -6.0)
  node[treatment label, anchor=east] {AutoBox1};
\draw[treatment line] ([yshift=-2pt] axis cs:15.452380952380953, 0) |- (axis cs:19.444555444555444, -5.0)
  node[treatment label, anchor=east] {AutoBox3};
\draw[treatment line] ([yshift=-2pt] axis cs:15.7995337995338, 0) |- (axis cs:19.444555444555444, -4.0)
  node[treatment label, anchor=east] {THETAsm};
\draw[treatment line] ([yshift=-2pt] axis cs:16.668498168498168, 0) |- (axis cs:19.444555444555444, -3.0)
  node[treatment label, anchor=east] {SINGLE};
\draw[treatment line] ([yshift=-2pt] axis cs:17.111222111222112, 0) |- (axis cs:19.444555444555444, -2.0)
  node[treatment label, anchor=east] {NAIVE2};
\draw[group line] (axis cs:15.263236763236764, -2.6666666666666665) -- (axis cs:15.7995337995338, -2.6666666666666665);
\draw[group line] (axis cs:14.451048951048952, -3.3333333333333335) -- (axis cs:15.452380952380953, -3.3333333333333335);
\draw[group line] (axis cs:14.22860472860473, -5.333333333333333) -- (axis cs:15.263236763236764, -5.333333333333333);
\draw[group line] (axis cs:12.783050283050283, -2.0) -- (axis cs:13.824009324009324, -2.0);
\draw[group line] (axis cs:12.783050283050283, -2.2) -- (axis cs:13.824009324009324, -2.2);
\draw[group line] (axis cs:13.335497835497835, -3.3333333333333335) -- (axis cs:14.417249417249417, -3.3333333333333335);
\draw[group line] (axis cs:13.642690642690642, -4.666666666666667) -- (axis cs:15.126207126207126, -4.666666666666667);
\draw[group line] (axis cs:13.642690642690642, -4.866666666666667) -- (axis cs:15.126207126207126, -4.866666666666667);

\end{axis}
\end{tikzpicture}
\caption{Results of the statistical testing in a critical difference diagram. The x-axis is the average rank across time series, in terms of MASE performance. Black bars show methods that are not statistically significantly different from each other. We see that \{L,S\}GT obtains the best average rank and is statistically significantly better than all comparison methods.} 
\label{fig:stattest}
\end{figure*}

Table~\ref{tab:overall} displays the performance of point forecasting for a number of forecasting algorithms on the M3 time series, averaged over all series belonging to a particular category. The non-seasonal LGT and seasonal SGT models proposed in this paper are run using the Stan sampler. We run LGT on the non-seasonal series, i.e., the yearly and other categories, and SGT on the seasonal series, i.e., quarterly and monthly series. We call this procedure \{L,S\}GT when reporting overall results. Furthermore, we report the best per-category/per-metric algorithms from the original M3 competition participants. The only papers that we are aware of that have claimed to achieve some level of improvement over the original M3 competition participants, without cross-learning, are baggedETS as proposed by \cite{Bergmeir2016Bagging} and the Multi Aggregation Prediction Algorithm (MAPA) presented in \cite{Kourentzes2014Improving}.
Thus, we also include the sMAPE and MASE scores for the best variants of bagged algorithms as reported in \cite{Bergmeir2016Bagging} for yearly, quarterly, and monthly series. That paper doesn't provide results for the \textit{Other} category though. 
Regarding MAPA, the sMAPE metric used by those authors is the same as Equation \ref{equ:smape}, whereas they use a different definition of the MASE, where those authors always use $s=1$ while differentiating in-sample data. However, their results reported of the benchmarks are quite different. For example, sMAPE of auto ETS of the yearly time series is reported as $18.7$, while we calculate it as $17.00$. An explanation could be changes/improvements in the default parameter configuration of auto ETS in the last years. Due to these discrepancies, we include results for MAPA of an experimental run that we perform, and which consistently shows better results than what is reported in the original paper. Furthermore, we also include runs of the standard auto ETS, auto ARIMA, and THETA techniques from the R \emph{forecast} package \citep{Hyndman2008Automatic}, and the mean of the auto ETS and auto ARIMA, which is denoted by ``hybrid'' in our table. Finally, we also run a version of baggedETS ourselves, by running the baggedETS function from the \emph{forecast} package with default parameters.

Table~\ref{tab:overall} shows that the LGT/SGT algorithms are superior to their competitors in terms of the sMAPE and MASE metrics in almost all cases, with the exception of sMAPE for monthly time series, for which they still outperform all participants in the original M3.

Figure~\ref{fig:stattest} shows a critical difference diagram with the results of statistical testing. Methods are ranked per series on all series of the M3 dataset according to their performance on MASE, and then statistical testing is applied. We see that \{L,S\}GT  outperforms with statistical significance all comparison methods, which are our own runs of benchmarks as before, together with all original participants in the M3, for which per-series forecasts are available so that they can be included in a per-series rank calculation.

\begin{table}[htb]
\centering
\begin{tabular}{lrr}
\hline
\textcolor{black}{Series} &
\textcolor{black}{Avg [\%]} &
\textcolor{black}{Max [\%]}\\\hline
\textcolor{black}{Yearly} &
\textcolor{black}{0.43} &
\textcolor{black}{50.1}\\
\textcolor{black}{Other} &
\textcolor{black}{0.11} &
\textcolor{black}{5.9}\\
\textcolor{black}{Monthly} &
\textcolor{black}{0.39} &
\textcolor{black}{89.6}\\
\textcolor{black}{Quarterly} &
\textcolor{black}{0.17} &
\textcolor{black}{9.1}\\
\textcolor{black}{All} &
\textcolor{black}{0.35} &
\textcolor{black}{89.6}\\\hline
\end{tabular}
\caption{Average and Maximum of absolute deviation in percent of forecasted values across $7$ runs.}
\label{tab:absdiff}
\end{table}

Finally, it is important to note that as the proposed LGT/SGT models are fitted with MCMC (i.e., sampling from the posterior), each run yields slightly different results. However, the standard deviations of metrics that are also shown in Table~\ref{tab:overall} (e.g., $0.0006$ for Monthly series' MASE), are comparatively small, demonstrating that the performance is quite stable.
Furthermore, Table~\ref{tab:absdiff} shows average and maximum absolute differences across all forecasted values in all series across seven runs of the method. We see that there are some cases where these deviations are large (the maxima). This happens when the true value is small. However, the average deviations are small, e.g., below $0.5\%$, so the method is usually stable across runs.

\begin{table*}[ht]
\centering\footnotesize
\begin{tabular}{lrrrr|rr}
  \hline
  \hline
 & Below 99p & Below 95p & Below 5p & Below 1p & MSIS 90p & MSIS 98p \\ 
 \hline
 \multicolumn{7}{c}{\centering \textit{All (amount of series: 3003)}}\\ 
    \hline
  \{L,S\}GT & \textbf{97.32} & \textbf{92.22} &  6.37 &  2.22 & \textbf{ 8.70} & \textbf{15.89} \\ 
  ETS/ZZZ & 95.87 & 90.75 & \textbf{ 5.35} & \textbf{ 1.79} &  9.79 & 20.10 \\ 
  ARIMA & 96.18 & 91.49 &  8.07 &  3.76 & 11.31 & 27.43 \\ 
  THETA & 95.37 & 89.87 &  6.28 &  2.50 & 10.09 & 22.02 \\ 
  MAPA & 94.02 & 88.30 &  7.03 &  3.18 & 10.98 & 26.95 \\   
   \hline
   \multicolumn{7}{c}{\centering \textit{Yearly (amount of series: 645, horizon: 6)}}\\ 
 \hline   
  LGT & \textbf{97.16} & \textbf{91.42} & \textbf{ 6.23} & \textbf{ 2.04} & \textbf{17.38} & \textbf{32.64} \\ 
  ETS/ZZZ & 91.81 & 86.41 &  7.26 &  3.88 & 21.80 & 50.49 \\ 
  ARIMA & 91.27 & 85.58 & 12.53 &  7.57 & 26.53 & 76.15 \\ 
  THETA & 91.47 & 84.34 &  \textbf{ 6.23} &  2.61 & 21.59 & 53.76 \\ 
  MAPA & 87.42 & 81.27 &  9.69 &  6.23 & 25.78 & 74.64 \\ 
 \hline
\multicolumn{7}{c}{\centering \textit{Other (amount of series: 174, horizon: 8)}}\\ 
    \hline
  LGT & 99.43 & 97.49 &  4.60 & \textbf{ 1.44} & \textbf{10.69} & \textbf{16.68} \\ 
  ETS/ZZZ & 98.20 & \textbf{94.54} &  6.11 &  2.16 & 10.93 & 18.02 \\ 
  ARIMA & \textbf{98.78} & 95.47 &  5.46 &  2.95 & 12.37 & 19.99 \\ 
  THETA & 99.57 & 97.56 &  9.55 &  3.16 & 13.32 & 20.64 \\ 
  MAPA & 98.28 & 95.69 & \textbf{ 5.32} &  1.94 & 11.75 & 17.12 \\ 
 \hline
\multicolumn{7}{c}{\centering \textit{Monthly (amount of series: 1428, horizon: 18)}}\\ 
    \hline
  SGT & \textbf{97.51} & \textbf{92.55} & \textbf{ 5.21} &  1.69 & \textbf{ 5.10} & \textbf{ 8.20} \\ 
  ETS/ZZZ & 96.55 & 91.37 &  4.28 & \textbf{ 1.12} &  5.17 &  8.42 \\ 
  ARIMA & 97.04 & 92.37 &  5.93 &  2.31 &  5.54 &  9.94 \\ 
  THETA & 95.81 & 90.35 &  5.69 &  2.26 &  5.62 & 10.25 \\ 
  MAPA & 94.95 & 88.99 &  5.57 &  2.09 &  5.50 & 10.23 \\ 
 \hline
\multicolumn{7}{c}{\centering \textit{Quarterly (amount of series: 756, horizon: 8)}}\\ 
    \hline
  SGT & \textbf{96.13} & 90.16 & 11.79 &  4.76 & \textbf{ 7.64} & \textbf{15.95} \\ 
  ETS/ZZZ & 95.02 & 90.01 &  8.50 & \textbf{ 3.22} &  7.99 & 16.71 \\ 
  ARIMA & 95.11 & \textbf{90.62} & 14.90 &  7.64 &  8.97 & 20.61 \\ 
  THETA & 95.04 & 89.60 & \textbf{ 8.09} &  3.29 &  7.98 & 17.47 \\ 
  MAPA & 93.35 & 88.18 & 11.94 &  6.15 &  8.52 & 20.08 \\ 
 \hline

\end{tabular}
\caption{Results for probabilistic forecast evaluation. Below 99/95/5/1p -- percentage of cases when the true value turned out to be below the 99th/95th/5th/1st percentile; MSIS 90p/98p -- MSIS for intervals of size that contain 90/98 percent of the observations, respectively.} 
\label{tab:percent}
\end{table*}

\subsection{Prediction Intervals}\label{sec:predint}

\begin{figure}[htb]
  \begin{center}
    \includegraphics[width=0.5\textwidth]{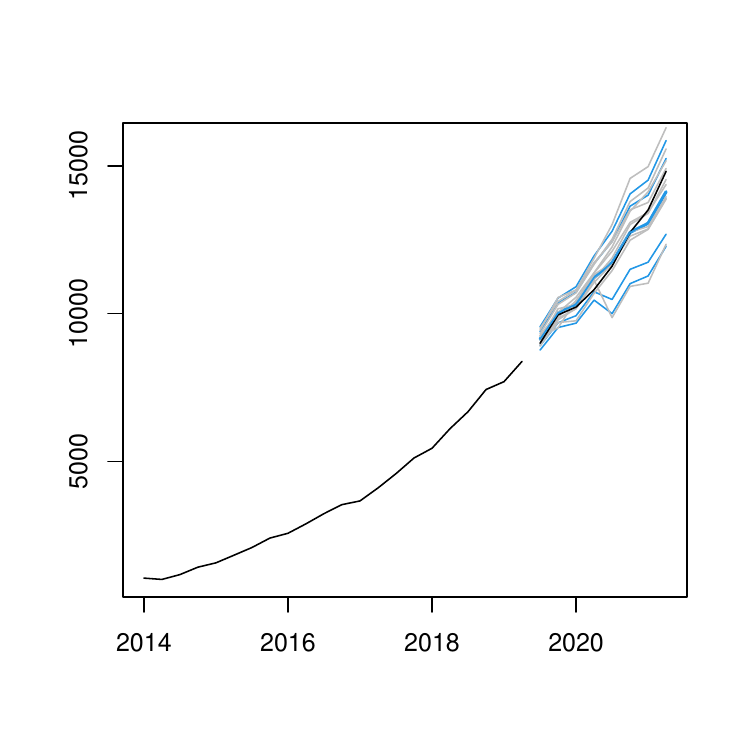}
    \caption{
    Quarterly revenue of Amazon Web Services (in million USD, source: statista.com). LGT forecast 2 years ahead, ten examples of the forecasting trajectories (in light gray), and the 1st, 5th, 50th, 95th, and 99th percentiles (in blue). Actuals shown in black.
    }
  \label{fig:example}
  \end{center}
\end{figure}

The forecasts are a result of step-by-step simulation with, e.g., $5000$ paths, and then aggregation by calculating quantiles. Figure~\ref{fig:example} shows a particular monthly time series, the first 10 of the simulation paths, the 1st, 5th, 95th, and 99th percentiles, and the 50th percentile taken as the expected value.

The prediction interval coverage appears to be category-dependent. Table~\ref{tab:percent} shows coverage ranges for our proposed methods, and analogous ranges of auto ETS, auto ARIMA, THETA, and MAPA with analytically calculated prediction intervals. Note that the original M3 participants are not compared here as they did not predict intervals, and also baggedETS is not readily able to produce prediction intervals. ETS and ARIMA are known to produce too narrow intervals, which can be confirmed from the table. While also our method tends to produce too narrow intervals (for example, for yearly series forecasted, in 97.16\% of cases the forecasted values are lower than 99p, but this should have happened in 99\% of cases), in the large majority of cases, the model performs better than the comparison methods. Interestingly, though not shown in this table, the simulated intervals of ETS and ARIMA are worse and even narrower. Checking the MSIS, which measures both calibration and sharpness of the intervals, we see that our proposed method consistently outperforms the comparison methods in all cases.

\subsection{Ablation study and computation times}
\label{sec:ablation}

\begin{table*}[ht!]
\centering
\begingroup\footnotesize
\begin{tabular}{lrrrrr}
  \hline
  \hline
 & sMAPE & MASE & MSIS 90p & MSIS 98p & Avg Runtime [s]\\ 
    \hline
   & \multicolumn{5}{c}{\centering \textit{All (amount of series: 3003)}}\\
   \hline
     ETS/ZZZ (forecast R package) &  13.07 &   1.43 &   9.79 &  20.10 &   0.30 \\ 
  ETS/\{AAN,MAM\} (forecast R package) &  14.41 &   1.52 &  11.66 &  28.17 &   0.04 \\ 
  ETS/BoxCox (forecast R package) &  13.28 &   1.53 &  10.81 &  23.86 &   0.01 \\ 
  \{L,S\}GT (our R package, Stan sampler) & \textbf{ 12.29} & \textbf{  1.30} & \textbf{  8.70} & \textbf{ 15.89} & 368.03 \\ 
  \{L,S\}GT without global trend &  13.04 &   1.45 &   9.34 &  16.07 & 271.47 \\ 
  \{L,S\}GT without heteroscedasticity &  12.29 &   1.31 &   9.54 &  19.81 & 421.87 \\ 
  \{L,S\}GT without student distribution &  12.41 &   1.31 &   8.74 &  16.11 & 298.19 \\ 
   \hline
   & \multicolumn{5}{c}{\centering \textit{Yearly (amount of series: 645, horizon: 6)}}\\
   \hline
  ETS/ZZZ (forecast R package) & 17.00 &  2.86 & 21.80 & 50.49 &  0.01 \\ 
  ETS/AAN (forecast R package) & 18.72 &  3.07 & 28.66 & 84.79 &  0.01 \\ 
  ETS/BoxCox (forecast R package) & 18.00 &  3.24 & 25.88 & 66.81 &  0.00 \\ 
  LGT (our R package, Stan sampler) & 15.18 & \textbf{ 2.48} & 17.38 & 32.64 & 50.40 \\ 
  LGT (our R package, Gibbs sampler) & \textbf{14.89} &  2.54 & 18.46 & 46.97 &  3.79 \\ 
  LGT (using MLE within forecast R package) & 21.49 &  3.29 & 27.77 & 73.49 &  0.01 \\ 
  LGT without global trend & 16.23 &  2.71 & 17.52 & \textbf{29.07} & 18.95 \\ 
  LGT without heteroscedasticity & 14.99 &  2.53 & 20.52 & 47.90 & 50.69 \\ 
  LGT without student distribution & 15.35 &  2.50 & \textbf{17.24} & 32.36 & 37.51 \\ 
  ETS/AAN (Bayesian version) & 15.94 &  2.81 & 21.76 & 48.85 & 14.33 \\ 
   \hline
   & \multicolumn{5}{c}{\centering \textit{Other (amount of series: 174, horizon: 8)}}\\
   \hline
  ETS/ZZZ (forecast R package) &   4.37 &   1.81 &  10.93 &  18.02 &   0.01 \\ 
  ETS/AAN (forecast R package) &   4.77 &   1.95 &  11.37 & \textbf{ 15.79} &   0.00 \\ 
  ETS/BoxCox (forecast R package) &   4.35 &   1.89 &  11.28 &  17.36 &   0.00 \\ 
  LGT (our R package, Stan sampler) &   4.25 &   1.72 & \textbf{ 10.69} &  16.68 & 161.51 \\ 
  LGT (our R package, Gibbs sampler) & \textbf{  4.23} & \textbf{  1.70} &  11.28 &  21.43 &   5.19 \\ 
  LGT (using MLE within forecast R package) &   5.01 &   2.13 &  15.82 &  22.86 &   0.02 \\ 
  LGT without global trend &   4.60 &   2.03 &  12.23 &  18.23 &  55.38 \\ 
  LGT without heteroscedasticity &   4.24 &   1.71 &  10.74 &  16.79 & 206.74 \\ 
  LGT without student distribution &   4.34 &   1.79 &  11.46 &  17.58 &  90.83 \\ 
  ETS/AAN (Bayesian version) &   4.71 &   2.09 &  12.96 &  19.97 &  39.48 \\ 
   \hline
   & \multicolumn{5}{c}{\centering \textit{Monthly (amount of series: 1428, horizon: 18)}}\\ 
   \hline
  ETS/ZZZ (forecast R package) &  14.14 &   0.86 &   5.17 &   8.42 &   0.58 \\ 
  ETS/MAM (forecast R package) &  15.54 &   0.92 &   5.64 &   9.37 &   0.08 \\ 
  ETS/BoxCox (forecast R package) &  14.17 &   0.87 &   5.32 &   8.76 &   0.03 \\ 
  SGT (our R package, Stan sampler) & \textbf{ 13.77} & \textbf{  0.83} & \textbf{  5.10} & \textbf{  8.20} & 444.97 \\ 
  SGT without global trend &  14.42 &   0.93 &   5.88 &   9.70 & 444.69 \\ 
  SGT without heteroscedasticity &  13.87 &   \textbf{  0.83} &   5.29 &   8.89 & 561.22 \\ 
  SGT without student distribution &  13.86 &   0.84 &   5.15 &   8.55 & 471.34 \\ 
   \hline
   & \multicolumn{5}{c}{\centering \textit{Quarterly (amount of series: 756, horizon: 8)}}\\
   \hline
  ETS/ZZZ (forecast R package) &   9.68 &   1.17 &   7.99 &  16.71 &   0.08 \\ 
  ETS/MAM (forecast R package) &  10.83 &   1.22 &   8.58 &  18.22 &   0.01 \\ 
  ETS/BoxCox (forecast R package) &   9.63 &   1.22 &   8.21 &  17.23 &   0.01 \\   
  SGT (our R package, Stan sampler) &   \textbf{  8.87} &   \textbf{  1.07} & \textbf{  7.64} & \textbf{ 15.95} & 541.21 \\ 
  SGT without global trend &   9.64 &   1.22 &   8.22 &  16.50 & 209.46 \\ 
  SGT without heteroscedasticity & \textbf{  8.87} & \textbf{  1.07} &   7.92 &  17.14 & 524.86 \\ 
  SGT without student distribution &   9.02 &   1.08 &   7.65 &  16.18 & 241.28 \\ 
   \hline
\end{tabular}
	\caption{Performance on M3 data of the ablation study methods. Average runtime is given in seconds.}
	\label{tab:ablation}
\endgroup
\end{table*}

While the LGT and SGT approaches are based on ETS methods, they differ from them in multiple ways. The main differences are concentrated between ETS (AAN) and LGT, while SGT can be considered as a direct extension of LGT into the space of seasonal data; the ETS model closest to SGT is the AAM model. 
Thus, in this ablation study, for non-seasonal data we investigate differences between ETS (AAN) and LGT to find the importance of the innovations and whether the complete set of these innovations is required to achieve the reported performance. Seasonal data is studied in a similar way.
We run performance testing on different variants of our proposed LGT and SGT approaches. 
The results of the ablation study are reported in Table~\ref{tab:ablation}.
We remove one by one features such as Student $t$-distribution and heteroscedasticity for errors and replace them with constant variance and normal distribution, respectively. Also, we remove the global trend feature. We note that SGT without global trend has then no trend at all, as SGT has no local trend. As overall comparison methods, we report results of a run of auto ETS (denoted ETS/ZZZ), and a run of auto ETS with a Box-Cox transformation as preprocessing. In the Box-Cox transformation, we choose the parameter $\lambda$ with the method from \cite{Guerrero1993Time}, which is also implemented in the \emph{forecast} package in R. As a deviation from the default behaviour of the implementation, we restrict $\lambda$ to be non-negative, as otherwise we observe an explosive behaviour on some time series and the method returns \emph{NA} values as forecasts.
Furthermore, for non-seasonal data we run our custom Gibbs sampler that is available for the non-seasonal case, and further run ETS/AAN, ETS/AAN (Bayesian version) -- ETS (AAN) fitted with Stan using our Bayesian approach, and an implementation of the LGT method within the \emph{forecast} package in R, which is thus fitted using the standard optimisation method which is used to fit ETS methods.
In terms of seasonal data, the \emph{forecast} package in R does not allow to fit ETS/AAM models due to numerical instabilities, so we fit MAM models instead, using multiplicative error instead of additive error, to compare with the ablation versions of SGT.

The results in Table~\ref{tab:ablation} allow us to make the following observations.
The model version without Student $t$-distribution is systematically worse than the full \{L,S\}GT model throughout all categories, on both sMAPE and MASE. The differences are rather small for MASE. MSIS results also show similar results that the full version is better than the version without Student $t$-distribution overall as well as in almost all particular cases (except MSIS 90p for yearly and quarterly).

Regarding the model without heteroskedasticity, in terms of prediction intervals, which is the main reason to include heteroskedasticity in the model, the model without heteroskedasticity is systematically worse in terms of MSIS than the full model. For point forecasting, the model is comparable, sometimes slightly better sometimes slightly worse.
Removing the global trend from the models leads to systematically worse results by large margins. Comparing Bayesian ETS to its MLE counterparts, we observe that our implementation of Bayesian ETS/AAN outperforms both ETS/ZZZ and ETS/AAN on all measures in the yearly category, but is generally worse than the MLE versions on the other category.

As a part of the ablation study we also recorded average execution times (per single time series and rounded to the $0.01$ second) for the investigated methods. We can see that the Bayesian based methods are orders of magnitude slower than the original exponential smoothing techniques. But with the unimpressive performance of our implementation of LGT in the \emph{forecast} package, the Bayesian model fitting seems necessary to leverage the performance gains of the proposed models. Also note that the bespoke Gibbs sampler runs about an order of magnitude faster than the Stan implementation, with very competitive results, often outperforming the Stan implementation.
We can furthermore observe that removing the Student $t$-distribution and the global trend both speed up the models considerably. The model without Student $t$-distribution is still relatively performant, in particular it is able to outperform ETS/ZZZ in all cases but monthly MSIS 98p. Thus, it can be a good choice for faster computation, though in the case of non-seasonal data, with full LGT with the bespoke Gibbs sampler, an even faster model is available.

\section{Conclusions}\label{sec:conclusions}

We have presented two related models, the seasonal SGT and the non-seasonal LGT model, that are derived from the AAZ exponential smoothing models. They are designed to perform well on both fast growing, and often irregular, time series found in fast-growing businesses such as cloud services or rideshare businesses, and on the M3 competition dataset, where they bested the top-performing algorithms in every category/metric pair, as well as auto ARIMA and auto ETS from the \emph{forecast} R package.
We believe that the very good forecasting performance both in terms of point- and probabilistic forecasting of the models on the M3 dataset comes mostly from their use of flexible, non-linear functions for the trend and the size of the error. Standard Exponential Smoothing models can only be either linear or exponential in their trend – these models can deliver both of these two extremes and anything in-between. Similarly, the function for size of the error bridges Exponential Smoothing models with additive and multiplicative errors. A Student's $t$-distribution for the error term is also more capable to fit to both smooth and fat-tailed time series. Probabilistic programming systems, including Stan, provide a convenient platform for development of innovative time series algorithms.

The main drawback of our new methods is that the Stan-based methods are orders of magnitude (typically over a thousand times) slower than the original ETS techniques. Though we have been able to overcome this problem to a certain extent by the implementation of a bespoke Gibbs sampler, shaving off one order of magnitude for the non-seasonal case, our methods are still at least hundreds of times slower than the classical ETS methods. However, in the new forecasting world where large scale forecasting is typically performed very competitively by global cross-series models (often Machine Learning models), our method can have an important place in the forecaster's toolbox to tackle forecasting of a small number of series with low granularity and without covariates, in a generally data-scarce environment.

\section*{Acknowledgements}

Christoph Bergmeir is supported by a María Zambrano (Senior) Fellowship that is funded by the Spanish Ministry of Universities and Next Generation funds from the European Union. He was also supported by the Australian Research Council under grant DE190100045 for part of the work. Erwin Wibowo undertook the work on the paper while affiliated with Monash University.  Alexander Dokumentov was funded by Monash University to undertake this work.

\bibliographystyle{agsm}

\bibliography{twomodels}

\end{document}